# Selecting Computations: Theory and Applications

**Nicholas Hay and Stuart Russell**
Computer Science Division
University of California
Berkeley, CA 94720
{nickjhay,russell}@cs.berkeley.edu

**David Tolpin and Solomon Eyal Shimony**
Department of Computer Science
Ben-Gurion University of the Negev
Beer Sheva 84105, Israel
{tolpin,shimony}@cs.bgu.ac.il

## Abstract

Sequential decision problems are often approximately solvable by simulating possible future action sequences. *Metalevel* decision procedures have been developed for selecting *which* action sequences to simulate, based on estimating the expected improvement in decision quality that would result from any particular simulation; an example is the recent work on using bandit algorithms to control Monte Carlo tree search in the game of Go. In this paper we develop a theoretical basis for metalevel decisions in the statistical framework of Bayesian *selection problems*, arguing (as others have done) that this is more appropriate than the bandit framework. We derive a number of basic results applicable to Monte Carlo selection problems, including the first finite sampling bounds for optimal policies in certain cases; we also provide a simple counterexample to the intuitive conjecture that an optimal policy will necessarily reach a decision in all cases. We then derive heuristic approximations in both Bayesian and distribution-free settings and demonstrate their superiority to bandit-based heuristics in one-shot decision problems and in Go.

## 1 Introduction

The broad family of sequential decision problems includes combinatorial search problems, game playing, robotic path planning, model-predictive control problems, Markov decision processes (MDP), whether fully or partially observable, and a huge range of applications. In almost all realistic instances, exact solution is intractable and approximate methods are sought. Perhaps the most popular approach is to simulate a limited number of possible future action sequences, in order to find a move in the current state that is (hopefully) near-optimal. In this paper, we develop a theoretical framework to examine the problem of selecting *which* future sequences to simulate. We derive a number of new results concerning optimal policies for this selection problem as well as new heuristic policies for controlling Monte Carlo simulations. As described below, these policies outperform previously published methods for "flat" selection and game-playing in Go.

The basic ideas behind our approach are best explained in a familiar context such as game playing. A typical game-playing algorithm chooses a move by first exploring a tree or graph of move sequences and then selecting the most promising move based on this exploration. Classical algorithms typically explore in a fixed order, imposing a limit on exploration depth and using pruning methods to avoid irrelevant subtrees; they may also reuse some previous computations (see Section 6.2). Exploring unpromising or highly predictable paths to great depth is often wasteful; for a given amount of exploration, decision quality can be improved by directing exploration towards those actions sequences whose outcomes are helpful in selecting a good move. Thus, the *metalevel* decision problem is to choose what future action sequences to explore (or, more generally, what deliberative computations to do), while the *object-level* decision problem is to choose an action to execute in the real world.

That the metalevel decision problem can itself be formulated and solved decision-theoretically was noted by Matheson (1968), borrowing from the related concept of *information value theory* (Howard, 1966). In essence, computations can be selected according to the expected improvement in decision quality resulting from their execution. I. J. Good (1968) independently proposed using this idea to control search in chess, and later defined "Type II rationality" to refer to agents that optimally solve the metalevel decision problem before acting. As interest in probabilistic and decision-theoretic approaches in AI grew during the 1980s, several authors explored these ideas further (Dean and Boddy, 1988; Doyle, 1988; Fehling and Breese, 1988; Horvitz, 1987). Work by Russell and Wefald (1988, 1991a,b) formulated the metalevel sequential decision problem, employing an explicit model of the results of computational actions, and applied this to the control of game-playing search in Othello with encouraging results.

An independent thread of research on metalevel control began with work by Kocsis and Szepesvári (2006) on the UCT algorithm, which operates in the context of *Monte Carlo tree search* (MCTS) algorithms. In MCTS, each computation takes the form of a simulation of a randomized sequence of actions leading from a leaf of the current tree to a terminal state. UCT is primarily a method for selecting a leaf from which to conduct the next simulation, and forms the core of the successful MoGo algorithm for Go playing (Gelly and Silver, 2011). The UCT algorithm is based on the the theory of bandit problems (Berry and Fristedt, 1985) and the asymptotically near-optimal UCB1 bandit algorithm (Auer et al., 2002). UCT applies UCB1 recursively to select actions to perform within simulations.

It is natural to consider whether the two independent threads are consistent; for example, are bandit algorithms such as UCB1 approximate solutions to some particular case of the metalevel decision problem defined by Russell and Wefald? The answer, perhaps surprisingly, is no. The essential difference is that, in bandit problems, every trial involves executing a real object-level action with real costs, whereas in the metareasoning problem the trials are *simulations* whose cost is usually independent of the utility of the action being simulated. Hence, as Audibert et al. (2010) and Bubeck et al. (2011) have also noted, UCT applies bandit algorithms to problems that are not bandit problems.

One consequence of the mismatch is that bandit policies are inappropriately biased away from exploring actions whose current utility estimates are low. Another consequence is the absence of any notion of "stopping" in bandit algorithms, which are designed for infinite sequences of trials. A metalevel policy needs to decide when to stop deliberating and execute a real action.

Analyzing the metalevel problem within an appropriate theoretical framework ought to lead to more effective algorithms than those obtained within the bandit framework. For Monte Carlo computations, in which samples are gathered to estimate the utilities of actions, the metalevel decision problem is an instance of the *selection problem* studied in statistics (Bechhofer, 1954; Swisher et al., 2003). Despite some recent work (Frazier and Powell, 2010; Tolpin and Shimony, 2012b), the theory of selection problems is less well understood than that of bandit problems. Most work has focused on the probability of selection error rather than optimal policies in the Bayesian setting (Bubeck et al., 2011). Accordingly, we present in Sections 2 and 3 a number of results concerning optimal policies for the general case as well as specific finite bounds on the number of samples collected by optimal policies for Bernoulli arms with beta priors. We also provide a simple counterexample to the intuitive conjecture that an optimal policy should not spend more on deciding than the decision is worth; in fact, it is possible for an optimal policy to compute forever. We also show by counterexample that optimal *index policies* (Gittins, 1989) may not exist for selection problems.

Motivated by this theoretical analysis, we propose in Sections 4 and 5 two families of heuristic approximations, one for the Bayesian case and one for the distribution-free setting. We show empirically that these rules give better performance than UCB1 on a wide range of standard (non-sequential) selection problems. Section 6 shows similar results for the case of guiding Monte Carlo tree search in the game of Go.

## 2 On optimal policies for selection

In a selection problem the decision maker is faced with a choice among alternative arms[1]. To make this choice, they may gather evidence about the utility of each of these alternatives, at some cost. The objective is to maximize the *net utility*, i.e., the expected utility of the final arm selected, less the cost of gathering the evidence. In the classical case (Bechhofer, 1954), evidence might consist of physical samples from a product batch; in a metalevel problem with Monte Carlo simulations, the evidence consists of outcomes of sampling computations:

**Definition 1.** *A **metalevel probability model** is a tuple $(U_1, \ldots, U_k, \mathcal{E})$ consisting of jointly distributed random variables:*

- *Real random variables $U_1, \ldots, U_k$, where $U_i$ is the utility of arm $i$, and*

- *A countable set $\mathcal{E}$ of random variables, each variable $E \in \mathcal{E}$ being a computation that can be performed and whose value is the result of that computation, where $e \in E$ will denote that $e$ is a potential value of the computation $E$.*

---

[1] Alternative actions are known as *arms* in the bandit setting; we borrow this terminology for uniformity.

For simplicity, in the below we'll assume the utilities $U_i$ are bounded, without loss of generality in $[0,1]$.

**Example 1** (Bernoulli sampling). *In the **Bernoulli metalevel probability model**, each arm will either succeed or not $U_i \in \{0,1\}$, with an unknown latent frequency of success $\Theta_i$, and a set of stochastic simulations of possible consequences $\mathcal{E} = \{E_{ij} | 1 \leq i \leq k, j \in \mathbb{N}\}$ that can be performed:*

$$\Theta_i \overset{iid}{\sim} \text{Uniform}[0,1] \quad \text{for } i \in \{1, \ldots, k\}$$
$$U_i \mid \Theta_i \sim \text{Bernoulli}(\Theta_i) \quad \text{for } i \in \{1, \ldots, k\}$$
$$E_{ij} \mid \Theta_i \overset{iid}{\sim} \text{Bernoulli}(\Theta_i) \quad \text{for } i \in \{1, \ldots, k\}, j \in \mathbb{N}$$

*The **one-armed Bernoulli metalevel probability model** has $k = 2$, $\Theta_1 = \lambda \in [0,1]$ a constant, and $\Theta_2 \sim \text{Uniform}[0,1]$.*

A metalevel probability model, when combined with a cost of computation $c > 0$,[2] defines a metalevel decision problem: what is the optimal strategy with which to choose a sequence of computations $E \in \mathcal{E}$ in order to maximize the agent's net utility? Intuitively, this strategy should choose the computations that give the most evidence relevant to deciding which arm to use, stopping when the cost of computation outweighs the benefit gained. We formalize the selection problem as a Markov Decision Process (see, e.g., Puterman (1994)):

**Definition 2.** *A (countable state, undiscounted) **Markov Decision Process** (MDP) is a tuple $M = (S, s_0, A_s, T, R)$ where: $S$ is a countable set of states, $s_0 \in S$ is the fixed initial state, $A_s$ is a countable set of actions available in state $s \in S$, $T(s, a, s')$ is the transition probability from $s \in S$ to $s' \in S$ after performing action $a \in A_s$, and $R(s, a, s')$ is the expected reward received on such a transition.*

To formulate the metalevel decision problem as an MDP, we define the states as sequences of computation outcomes and allow for a terminal state when the agent chooses to stop computing and act:

**Definition 3.** *Given a metalevel probability model*[3] *$(U_1, \ldots, U_k, \mathcal{E})$ and a cost of computation $c > 0$, a corresponding **metalevel decision problem** is any*

---

[2]The assumption of a fixed cost of computation is a simplification; precise conditions for its validity are given by Harada (1997).

[3]Definition 1 made no assumption about the computational result variables $E_i \in \mathcal{E}$, but for simplicity in the following we'll assume that each $E_i$ takes one of a countable set of values. Without loss of generality, we'll further assume the domains of the computational variables $E \in \mathcal{E}$ are disjoint.

*MDP $M = (S, s_0, A_s, T, R)$ such that*

$$S = \{\bot\} \cup \{\langle e_1 \ldots, e_n \rangle : e_i \in E_i \text{ for all } i,$$
$$\text{for finite } n \geq 0 \text{ and distinct } E_i \in \mathcal{E}\}$$
$$s_0 = \langle \rangle$$
$$A_s = \{\bot\} \cup \mathcal{E}_s$$

*where $\bot \in S$ is the unique terminal state, where $\mathcal{E}_s \subseteq \mathcal{E}$ is a state-dependent subset of allowed computations, and when given any $s = \langle e_1, \ldots, e_n \rangle \in S$, computational action $E \in \mathcal{E}$, and $s' = \langle e_1, \ldots, e_n, e \rangle \in S$ where $e \in E$, we have:*

$$T(s, E, s') = P(E = e \mid E_1 = e_1, \ldots, E_n = e_n)$$
$$T(s, \bot, \bot) = 1$$
$$R(s, E, s') = -c$$
$$R(s, \bot, \bot) = \max_i \mu_i(s)$$

*where $\mu_i(s) = \mathbb{E}[U_i \mid E_1 = e_1, \ldots, E_n = e_n]$.*

Note that when stopping in state $s$, the expected utility of action $i$ is by definition $\mu_i(s)$, so the optimal action to take is $i^* \in \arg\max_i \mu_i(s)$ which has expected utility $\mu_{i*}(s) = \max_i \mu_i(s)$.

One can optionally add an external constraint on the number of computational actions, or their total cost, in the form of a deadline or *budget*. This bridges with the related area of budgeted learning (Madani et al., 2004). Although this feature is not formalized in the MDP, it can be added by including either time or past total cost as part of the state.

**Example 2** (Bernoulli sampling). *In the Bernoulli metalevel probability model (Example 1), note that:*

$$\Theta_i \mid E_{i1}, \ldots, E_{in_i} \sim \text{Beta}(s_i + 1, f_i + 1) \quad (1)$$
$$E_{i(n_i+1)} \mid E_{i1}, \ldots, E_{in_i} \sim \text{Bernoulli}\left(\frac{s_i + 1}{n_i + 2}\right) \quad (2)$$
$$\mathbb{E}[U_i \mid E_{i1}, \ldots, E_{in_i}] = (s_i + 1)/(n_i + 2) \quad (3)$$

*by standard properties of these distributions, where $s_i = \sum_{j=1}^{n_i} E_{in_i}$ is the number of simulated successes of arm $i$, and $f_i = n_i - s_i$ the failures. By Equation (1), the state space is the set of all $k$ pairs $(s_i, f_i)$; Equations (2) and (3) suffice to give the transition probabilities and terminal rewards, respectively. The one-armed Bernoulli case is similar, requiring as state just $(s, f)$ defining the posterior over $\Theta_2$.*

Given a metalevel decision problem $M = (S, s_0, A_s, T, R)$ one defines policies and value functions as in any MDP. A (deterministic, stationary) **metalevel policy** $\pi$ is a function mapping states $s \in S$ to actions to take in that state $\pi(s) \in A_s$.

The **value function** for a policy $\pi$ gives the expected total reward received under that policy starting from a given state $s \in S$, and the **Q-function** does the same when starting in a state $s \in S$ and taking a given action $a \in A_s$:

$$V_M^\pi(s) = \mathbb{E}_M^\pi \left[ \sum_{i=0}^N R(S_i, \pi(S_i), S_{i+1}) \mid S_0 = s \right] \quad (4)$$

where $N \in [0, \infty]$ is the random time the MDP is terminated, i.e., the unique time where $\pi(S_N) = \bot$, and similarly for the Q-function $Q_M^\pi(s, a)$.

As usual, an **optimal policy** $\pi^*$, when it exists, is one that maximizes the value from every state $s \in S$, i.e., if we define for each $s \in S$

$$V_M^*(s) = \sup_\pi V_M^\pi(s),$$

then an optimal policy $\pi^*$ satisfies $V_M^{\pi^*}(s) = V_M^*(s)$ for all $s \in S$, where we break ties in favor of stopping.

The optimal policy must balance the cost of computations with the improved decision quality that results. This tradeoff is made clear in the value function:

**Theorem 4.** *The value function of a metalevel decision process* $M = (S, s_0, A_s, T, R)$ *is of the form*

$$V_M^\pi(s) = \mathbb{E}_M^\pi[-c N + \max_i \mu_i(S_N) \mid S_0 = s]$$

*where $N$ denotes the (random) total number of computations performed; similarly for $Q_M^\pi(s, a)$.*

In many problems, including the Bernoulli sampling model of Example 2, the state space is infinite. Does this preclude solving for the optimal policy? Can infinitely many computations be performed?

There is in full generality an upper bound on the *expected* number of computations a policy performs:

**Theorem 5.** *The optimal policy's expected number of computations is bounded by the value of perfect information (Howard, 1966) times the inverse cost $1/c$:*

$$\mathbb{E}^{\pi^*}[N \mid S_0 = s] \leq \frac{1}{c} \left( \mathbb{E}[\max_i U_i \mid S_0 = s] - \max_i \mu_i(s) \right).$$

*Further, any policy $\pi$ with infinite expected number of computations has negative infinite value, hence the optimal policy stops with probability one.*

Although the *expected* number of computations is always bounded, there are important cases in which the *actual* number is not, such as the following inspired by the sequential probability ratio test (Wald, 1945):

**Example 3.** *Consider the Bernoulli sampling model for two arms but with a different prior:* $\Theta_1 = 1/2$, *and $\Theta_2$ is 1/3 or 2/3 with equal probability. Simulating arm 1 gains nothing, and after $(s, f)$ simulated successes and failures of arm 2 the posterior odds ratio is*

$$\frac{P(\Theta_2 = 2/3 \mid s, f)}{P(\Theta_2 = 1/3 \mid s, f)} = \frac{(2/3)^s (1/3)^f}{(1/3)^s (2/3)^f} = 2^{s-f}.$$

*Note that this ratio completely specifies the posterior distribution of $\Theta_2$, and hence the distribution of the utilities and all future computations. Thus, whether it is optimal to continue is a function only of this ratio, and thus of $s - f$. For sufficiently low cost, the optimal policy samples when $s - f$ equals $-1$, 0, or 1. But with probability 1/3, a state with $s - f = 0$ transitions to another state $s - f = 0$ after two samples, giving finite, although exponentially decreasing, probability to arbitrarily long sequences of computations.*

However, in a number of settings, including the original Bernoulli model of Example 1, we can prove an upper bound on the number of computations. For reasons of space, and for its later use in Section 4, we prove here the bound for the one-armed Bernoulli model.

Before we can do this, we need to get an analytical handle on the optimal policy. The key is through a natural approximate policy:

**Definition 6.** *Given a metalevel decision problem* $M = (S, s_0, A_s, T, R)$, *the **myopic policy** $\pi^m(s)$ is defined to equal $\arg\max_{a \in A_s} Q^m(s, a)$ where $Q^m(s, \bot) = \max_i \mu_i(s)$ and*

$$Q^m(s, E) = \mathbb{E}_M[-c + \max_i \mu_i(S_1) \mid S_0 = s, A_0 = E].$$

The myopic policy (known as the metalevel greedy approximation with single-step assumption in (Russell and Wefald, 1991a)) takes the best action, to either stop or perform a computation, under the assumption that at most one further computation can be performed. It has a tendency to stop too early, because changing one's mind about which real action to take often takes more than one computation. In fact, we have:

**Theorem 7.** *Given a metalevel decision problem $M = (S, s_0, A_s, T, R)$ if the myopic policy performs some computation in state $s \in S$, then the optimal policy does too, i.e., if $\pi^m(s) \neq \bot$ then $\pi^*(s) \neq \bot$.*

Despite this property, the stopping behavior of the myopic policy does have a close connection to that of the optimal policy:

**Definition 8.** *Given a metalevel decision problem* $M = (S, s_0, A_s, T, R)$, *a subset $S' \subseteq S$ of states is **closed under transitions** if whenever $s' \in S'$, $a \in A_{s'}$, $s'' \in S$, and $T(s', a, s'') > 0$, we have $s'' \in S'$.*

**Theorem 9.** *Given a metalevel decision problem $M = (S, s_0, A_s, T, R)$ and a subset $S' \subseteq S$ of states closed under transitions, if the myopic policy stops in all states $s' \in S'$ then the optimal policy does too.*

Using these results connecting the behavior of the optimal and myopic policies, we can prove our bound:

**Theorem 10.** *The one-armed Bernoulli decision process with constant arm $\lambda \in [0,1]$ performs at most $\lambda(1-\lambda)/c - 3 \leq 1/4c - 3$ computations.*

*Proof.* Using Definition 6 and Example 2, we determine which states the myopic policy stops in by bounding $Q^m(s, E)$. For a state $(s, f)$, let $\mu = (s+1)/(n+2)$ be the mean utility for arm 2, where $n = s + f$. Fixing $n$ and maximizing over $\mu$, we get sufficient condition for stopping Since the set of states satisfying Equation (5) is closed under

$$c \geq \frac{\lambda(1-\lambda)}{(n+3)} \qquad n \geq \frac{\lambda(1-\lambda)}{c} - 3 \qquad (5)$$

Since the set of states satisfying Equation (5) is closed under transitions ($n$ only increases), by Theorem 7. Finally, note $\max_{\lambda \in [0,1]} \lambda(1-\lambda) = 1/4$. □

A key implication is that the *optimal* policy can be computed in time $O(1/c^2)$, i.e., quadratic in the inverse cost. This is particularly appropriate when the cost of computation is relatively high, such as in simulation experiments (Swisher et al., 2003), or when the decision to be made is critical.

## 3 Context effects and non-indexability

The Gittins index theorem (Gittins, 1979) is a famous structural result for bandit problems. It states that in bandit problems with independent reward distribution for each arm and geometric discounting, the optimal policy is an **index policy**: each arm is assigned a real-valued index based on its state only, such that it is optimal to sample the arm with greatest index.

The analogous result does *not* hold for metalevel decision problems, even when the action's values are independent (this formalized later in Definition 13):

**Example 4** (Non-indexability). *Consider a metalevel probability model with three actions. $U_1$ is equally likely to be $-1.5$ or $1.5$ (low mean, high variance), $U_2$ is equally likely to be $0.25$ or $1.75$ (high mean, low variance), and $U_3 = \lambda$ has a known value (the context). The two computations are to observe exactly $U_1$ and $U_2$, respectively, each with cost $0.2$. The corresponding metalevel MDP has 9 states and can be solved exactly. Figure 1 plots $Q^*_\lambda(s_0, U_i) - Q^*_\lambda(s_0, \bot)$ as a function of the known value $\lambda$. As the context $\lambda$ varies the optimal action inverts from observing 1 to observing 2. Inversions like this are impossible for index policies.*

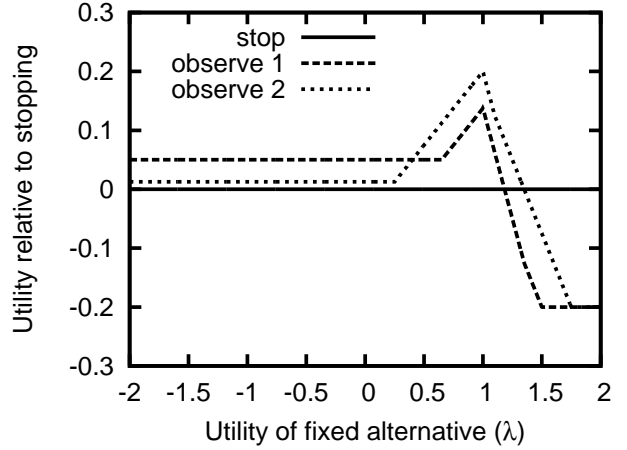

Figure 1: Optimal Q-values of computing relative to stopping as a function of the utility of the fixed alternative. Note the inversion where for low $\lambda$ observing action 1 is strictly optimal, while for medium $\lambda$ observing action 2 is strictly optimal.

There is, however, a restriction on what kind of influence the context can have:

**Definition 11.** *Given a metalevel decision problem $M = (S, s_0, A_s, T, R)$, and a constant $\lambda \in \mathbb{R}$, define $M_\lambda = (S, s_0, A_s, T, R_\lambda)$ to be $M$ with an additional action of known value $\lambda$, defined by:*

$$R_\lambda(s, E, s') = R(s, E, s')$$
$$R_\lambda(s, \bot, \bot) = \max\{\lambda, R(s, \bot, \bot)\}$$

Note this is equivalent to adding an extra arm with constant value $U_{k+1} = \lambda$.

**Theorem 12.** *Given a metalevel decision problem $M = (S, s_0, A_s, T, R)$, there exists a real interval $I(s)$ for every state $s \in S$ such that it is optimal to stop in state $s$ in $M_\mu$ iff $\mu \notin I(s)$. Furthermore, $I(s)$ contains $\max_i \mu_i(s)$ whenever it is nonempty.*

## 4 The blinkered policy

The myopic policy is an extreme approximation, often stopping far too early. A better approximation can be obtained, at least for the case where each computation can only affect the value of one action. The technical definition (closely related to *subtree independence* in Russell and Wefald's work) is as follows:

**Definition 13.** *A metalevel probability model $(U_1, \ldots, U_k, \mathcal{E})$ has **independent actions** if the computational variables can be partitioned $\mathcal{E} = \mathcal{E}_1 \cup \cdots \cup \mathcal{E}_k$*

*such that such that the sets $\{U_i\} \cup \mathcal{E}_i$ are independent of each other for different $i$.*

With independent actions, we can talk about metalevel policies that focus on computations affecting a single action. These policies are not myopic—they can consider arbitrarily many computations—but they are *blinkered* because they can look in only a single direction at a time:

**Definition 14.** *Given a metalevel decision problem $M = (S, s_0, A_s, T, R)$ with independent actions, the **blinkered policy** $\pi^b$ is defined by $\pi^b(s) = \operatorname{argmax}_{a \in A_s} Q^b(s, a)$ where $Q^b(s, \bot) = \bot$ and for $E_i \in \mathcal{E}_i$*

$$Q^b(s, E_i) = \sup_{\pi \in \Pi_i^b} Q^\pi(s, E_i) \qquad (6)$$

*where $\Pi_i^b$ is the set of policies $\pi$ where $\pi(s) \in \mathcal{E}_i$ for all $s \in S$.*

Clearly, blinkered policies are better than myopic: $Q^m(s, a) \leq Q^b(s, a) \leq Q^*(s, a)$. Moreover, the blinkered policy can be computed in time proportional to the number of arms, by breaking the decision problem into separate subproblems:

**Definition 15.** *Given a metalevel decision problem $M = (S, s_0, A_s, T, R)$ with independent actions, a **one-action metalevel decision problem** for $i = 1, \ldots, k$ is the metalevel decision problem $M^1_{i,\lambda} = (S_i, s_0, A_{s0}, T_i, R_i)$ defined by the metalevel probability model $(U_0, U_i, \mathcal{E}_i)$ with $U_0 = \lambda$.*

Note that given a state $s$ of a metalevel decision problem, we can form a state $s_i$ by taking only the results of computations in $\mathcal{E}_i$ (see Definition 3). By action independence, $\mu_i(s)$ is a function only of $s_i$.

**Theorem 16.** *Given a metalevel decision problem $M = (S, s_0, A_s, T, R)$ with independent actions, let $M^1_{i,\lambda_i}$ be the $i$th one-action metalevel decision problem for $i = 1, \ldots, k$. Then for any $s \in S$, whenever $E_i \in A_s \cap \mathcal{E}_i$ we have:*

$$Q^b_M(s, E_i) = Q^*_{M^1_{i,\mu^*_{-i}}}(s_i, E_i)$$

*where $\mu^*_{-i} = \max_{j \neq i} \mu_j(s)$.*

Theorem 16 shows that to compute the blinkered policy we need only compute the optimal policies for $k$ separate one-action problems.

For the Bernoulli problem with $k$ actions, the one-action metalevel decision problems are all one-action Bernoulli problems (Example 1). By Theorem 10 these policies perform at most $1/4c - 3$ computations. As a result, the blinkered policy can be numerically computed in time $O(D/c^2)$ independent of $k$ by backwards induction, where $D$ is the number of points $\lambda \in [0, 1]$ for which we compute $Q^*_{M^1_{i,m}}(s)$.[4] This will be worth the cost in the same situations as mentioned at the end of Section 2.

Figure 2 compares the blinkered policy to several other policies from the literature, using a Bernoulli sampling problem with $k = 25$ and a wide range of values for the step cost $c$. Performance is measured by expected *regret*, where the regret includes the cost of sampling: $R = (\max_i U_i) - U_j + cn$ where $n$ is the number of computations and $j$ is the action actually selected. The blinkered policy significantly outperforms all others. The myopic policy plateaus as it quickly reaches a position where no single computation can change the final action choice. ESPb performs quite well given that is making a normal approximation to the Beta posterior. The curves for UCB1-B and UCB1-b show that even given a good stopping rule, UCB1's choice of actions to sample is not ideal.

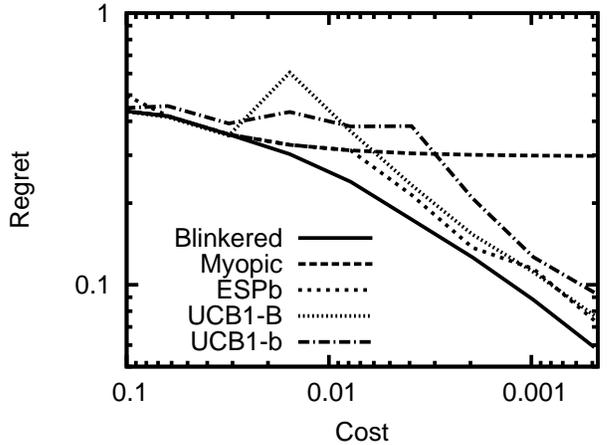

Figure 2: Average regret of various policies as a function of the cost in a 25-action Bernoulli sampling problem, over 1000 trials. Error bars omitted as they are negligible (the relative error is at most 0.03).

## 5 Upper bounds on Value of Information

In many practical applications of the selection problem, such as search in the game of Go, prior distributions are unavailable.[5] In such cases, one can still bound the value of information of myopic policies using *concentration inequalities* to derive distribution-independent bounds on the VOI. We obtain such bounds under the following assumptions:

---

[4]In our experiments below, $D = 129$ points are equally spaced, using linear interpolation between points.

[5]The analysis is also applicable to some Bayesian settings, using "fake" samples to simulate prior distributions.

1. Samples are iid given the value of the arms (variables), as in the Bayesian schemes such as Bernoulli sampling.

2. The expectation of a selection in a belief state is equal to the sample mean (and therefore, after sampling terminates, the arm with the greatest sample mean will be selected).

When considering possible samples in the blinkered semi-myopic setting, two cases are possible: either the arm $\alpha$ with the highest sample mean $\overline{X}_\alpha$ is tested, and $\overline{X}_\alpha$ becomes lower than $\overline{X}_\beta$ of the second-best arm $\beta$; or, another arm $i$ is tested, and $\overline{X}_i$ becomes higher than $\overline{X}_\alpha$.

Our bounds below are applicable to any bounded distribution (without loss of generality bounded in $[0,1]$). Similar bounds can be derived for certain unbounded distributions, such as the normally distributed prior value with normally distributed sampling. We derive a VOI bound for testing an arm a fixed $N$ times, where $N$ can be the remaining budget of available samples or any other integer quantity. Denote by $\Lambda_i^b$ the intrinsic VOI of testing the $i$th arm $N$ times, and the number of samples already taken from the $i$th arm by $n_i$.

**Theorem 17.** $\Lambda_i^b$ is bounded from above as

$$\Lambda_\alpha^b \leq \frac{N\overline{X}_\beta^{n_\beta}}{n_\alpha} \Pr(\overline{X}_\alpha^{n_\alpha+N} \leq \overline{X}_\beta^{n_\beta})$$

$$\Lambda_{i|i\neq\alpha}^b \leq \frac{N(1-\overline{X}_\alpha^{n_\alpha})}{n_i} \Pr(\overline{X}_i^{n_i+N} \geq \overline{X}_\alpha^{n_\alpha}) \quad (7)$$

The probabilities can be bounded from above using the Hoeffding inequality (Hoeffding, 1963):

**Theorem 18.** The probabilities in Equation (7) are bounded from above as

$$\Pr(\overline{X}_\alpha^{n_\alpha+N} \leq \overline{X}_\beta^{n_\beta}) \leq 2\exp\left(-\varphi(\overline{X}_\alpha^{n_\alpha} - \overline{X}_\beta^{n_\beta})^2 n_\alpha\right)$$

$$\Pr(\overline{X}_{i|i\neq\alpha}^{n_\alpha+N} \geq \overline{X}_\beta^{n_\beta}) \leq 2\exp\left(-\varphi(\overline{X}_\alpha^{n_\alpha} - \overline{X}_i^{n_i})^2 n_i\right) \quad (8)$$

where $\varphi = \min\left(2(\frac{1+n/N}{1+\sqrt{n/N}})^2\right) = 8(\sqrt{2}-1)^2 > 1.37$.

**Corollary 19.** An upper bound on the VOI estimate $\Lambda_i^b$ is obtained by substituting Equation (8) into (7).

$$\Lambda_\alpha^b \leq \hat{\Lambda}_\alpha^b = \frac{2N\overline{X}_\beta^{n_\beta}}{n_\alpha} \exp\left(-\varphi(\overline{X}_\alpha^{n_\alpha} - \overline{X}_\beta^{n_\beta})^2 n_\alpha\right) \quad (9)$$

$$\Lambda_{i|i\neq\alpha}^b \leq \hat{\Lambda}_i^b = \frac{2N(1-\overline{X}_\alpha^{n_\alpha})}{n_i} \exp\left(-\varphi(\overline{X}_\alpha^{n_\alpha} - \overline{X}_i^{n_i})^2 n_i\right)$$

More refined bounds can be obtained through tighter estimates on the probabilities in Equation (7), for example, based on the empirical Bernstein inequality (Maurer and Pontil, 2009), or through a more careful application of the Hoeffding inequality, resulting in:

$$\Lambda_i^b \leq \frac{N\sqrt{\pi}}{n_i\sqrt{n_i}}\left[\mathrm{erf}\left((1-\overline{X}_i^{n_i})\sqrt{n_i}\right) - \mathrm{erf}\left((\overline{X}_\alpha^{n_\alpha} - \overline{X}_i^{n_i})\sqrt{n_i}\right)\right]$$

$$\Lambda_\alpha^b \leq \frac{N\sqrt{\pi}}{n_\alpha\sqrt{n_\alpha}}\left[\mathrm{erf}\left(\overline{X}_\alpha^{n_\alpha}\sqrt{n_\alpha}\right) - \mathrm{erf}\left((\overline{X}_\alpha^{n_\alpha} - \overline{X}_\beta^{n_\beta})\sqrt{n_\alpha}\right)\right]$$
(10)

Selection problems usually separate out the decision of *whether* to sample or to stop (called the stopping policy), and *what* to sample. We'll examine the first issue here, along with the empirical evaluation of the above approximate algorithms, and the second in the following section.

Assuming that the sample costs are constant, a semi-myopic policy will decide to test the arm that has the best current VOI estimate. When the distributions are unknown, it makes sense to use the upper bounds established in Theorem 17, as we do in the following. This evaluation assumes a fixed budget of samples, which is completely used up by each of the candidate schemes, making a stopping criterion irrelevant.

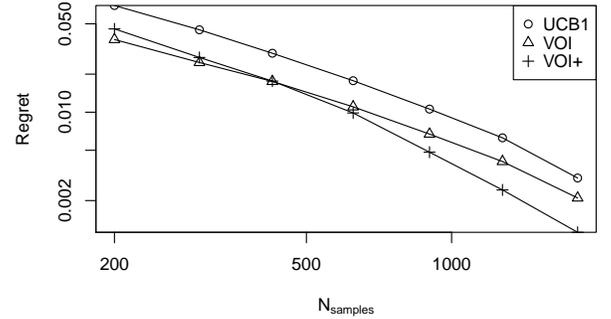

Figure 3: Average regret of various policies as a function of the fixed number of samples in a 25-action Bernoulli sampling problem, over 10000 trials.

The sampling policies are compared on random Bernoulli selection problem instances. Figure 3 shows results for randomly-generated selection problems with 25 Bernoulli arms, where the mean rewards of the arms are distributed uniformly in $[0,1]$, for a range of sample budgets 200..2000, with multiplicative step of 2, averaging over 10000 trials. We compare UCB1 with the policies based on the bounds in Equation (9) (VOI) and Equation (10) (VOI+). UCB1 is always considerably worse than the VOI-aware sampling policies.

## 6 Sampling in trees

The previous section addressed the selection problem in the flat case. Selection in trees is more complicated. The goal of Monte-Carlo tree search (Chaslot et al., 2008) at the root node is usually to select an action that appears to be the best based on outcomes of *search rollouts*. But the goal of rollouts at non-root nodes is different than at the root: here it is important to better approximate the value of the node, so that selection at the root can be more informed. The exact analysis of sampling at internal nodes is outside the scope of this paper. At present we have no better proposal for internal nodes than to use UCT there.

We thus propose the following hybrid sampling scheme (Tolpin and Shimony, 2012a): at the *root node*, sample based on the VOI estimate; at *non-root nodes*, sample using UCT.

Strictly speaking, even at the root node the stationarity assumptions[6] underlying our belief-state MDP for selection do not hold exactly. UCT is an adaptive scheme, and therefore the values generated by sampling at non-root nodes will typically cause values observed at children of the root node to be non-stationary. Nevertheless, sampling based on VOI estimates computed as for stationary distributions works well in practice. As illustrated by the empirical evaluation (Section 6), estimates based on upper bounds on the VOI result in good sampling policies, which exhibit performance comparable to the performance of some state-of-the-art heuristic algorithms.

### 6.1 Stopping criterion

When a sample has a known cost commensurable with the value of information of a measurement, an upper bound on the intrinsic VOI can also be used to stop the sampling if the intrinsic VOI of any arm is less than the total cost of sampling $C$: $\max_i \Lambda_i \leq C$.

The VOI estimates of Equations (7) and (9) include the remaining sample budget $N$ as a factor, but given the cost of a single sample $c$, the cost of the remaining samples accounted for in estimating the intrinsic VOI is $C = cN$. $N$ can be dropped on both sides of the inequality, giving a reasonable stopping criterion:

$$\frac{1}{N}\Lambda_\alpha^b \leq \frac{\overline{X}_\beta^{n_\beta}}{n_\alpha} \Pr(\overline{X}_\alpha^{n_\alpha+N} \leq \overline{X}_\beta^{n_\alpha}) \leq c$$
$$\frac{1}{N}\max_i \Lambda_i^b \leq \max_i \frac{(1-\overline{X}_\alpha^{n_\alpha})}{n_i} \Pr(\overline{X}_i^{n_i+N} \geq \overline{X}_\alpha^{n_\alpha}) \leq c$$
$$\forall i : i \neq \alpha \qquad (11)$$

The empirical evaluation (Section 6) confirms the viability of this stopping criterion and illustrates the influence of the sample cost $c$ on the performance of the sampling policy. When the sample cost $c$ is unknown, one can perform initial calibration experiments to determine a reasonable value, as done in the following.

---

[6]This is not a restriction, however, of the general formalism in Section 2.

### 6.2 Sample redistribution in trees

The above hybrid approach assumes that the information obtained from rollouts in the current state is discarded after an real-world action is selected. In practice, many successful Monte-Carlo tree search algorithms reuse rollouts generated at earlier search states, if the sample traverses the current search state during the rollout; thus, the value of information of a rollout is determined not just by the influence on the choice of the action at the current state, but also by its potential influence on the choice at future search states.

One way to account for this reuse would be to incorporate the 'future' value of information into a VOI estimate. However, this approach requires a nontrivial extension of the theory of metareasoning for search. Alternately, one can behave myopically with respect to the search tree depth:

1. Estimate VOI as though the information is discarded after each step,

2. Stop early if the VOI is below a certain threshold (see Section 6.1), and

3. Save the unused sample budget for search in future states, such that if the nominal budget is $N$, and the unused budget in the last state is $N_u$, the search budget in the next state will be $N + N_u$.

In this approach, the cost $c$ of a sample in the current state is the VOI of increasing the budget of a future state by one sample. It is unclear whether this cost can be accurately estimated, but supposing a fixed value for a given problem type and algorithm implementation would work. Indeed, the empirical evaluation (Section 6.3) confirms that stopping and sample redistribution based on a learned fixed cost substantially improve the performance of the VOI-based sampling policy in game tree search.

### 6.3 Playing Go against UCT

The hybrid policies were compared on the game Go, a search domain in which UCT-based MCTS has been particularly successful (Gelly and Wang, 2006). A modified version of Pachi (Braudiš and Loup Gailly, 2011), a state of the art Go program, was used for the experiments:

- The UCT engine of Pachi was extended with VOI-aware sampling policies at the first step.

- The stopping criterion for the VOI-aware policy was modified and based solely on the sample cost, specified as a constant parameter. The heuristic stopping criterion for the original UCT policy was left unchanged.

- The time-allocation model based on the fixed number of samples was modified for *both the original UCT policy and the VOI-aware policies* such that
    - Initially, the same number of samples is available to the agent at each step, independently of the number of pre-simulated games;
    - If samples were unused at the current step, they become available at the next step.

While the UCT engine is not the most powerful engine of Pachi, it is still a strong player. On the other hand, additional features of more advanced engines would obstruct the MCTS phenomena which are the subject of the experiment. The engines were com-pared on the 9x9 board, for 5000, 7000, 1000, and 15000 samples (game simulations) per ply, each experiment repeated 1000 times. Figure 4 depicts a calibration experiment, showing the winning rate of the VOI-aware policy against UCT as a function of the stopping threshold $c$ (if the maximum VOI of a sample is below the threshold, the simulation is stopped, and a move is chosen). Each curve in the figure corresponds to a certain number of samples per ply. For the stopping threshold of $10^{-6}$, the VOI-aware policy is almost always better than UCT, and reaches the winning rate of 64% for 10000 samples per ply.

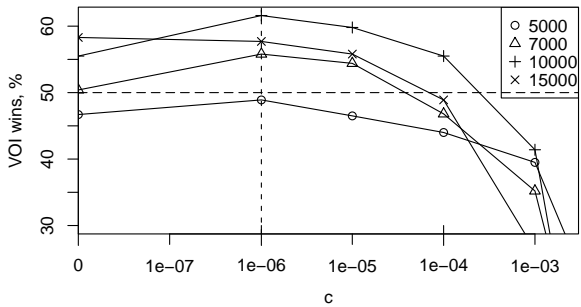

Figure 4: Winning rate of the VOI-aware policy in Go as a function of the cost $c$, for varying numbers of samples per ply.

Figure 5 shows the winning rate of VOI against UCT $c = 10^{-6}$. In agreement with the intuition (Figure 6.2), VOI-based stopping and sample redistribution is most influential for intermediate numbers of samples per ply. When the maximum number of samples is too low, early stopping would result in poorly selected moves. On the other hand, when the maximum number of samples is sufficiently high, the VOI of increasing the maximum number of samples in a future state is low.

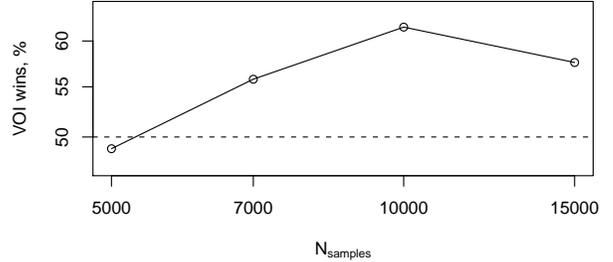

Figure 5: Winning rate of the VOI-aware policy in Go as a function of the number of samples, fixing cost $c = 10^{-6}$.

Note that if we disallowed reuse of samples in both Pachi and in our VOI-based scheme, the VOI based-scheme win rate is even higher than shown in Figure 5. This is as expected, as this setting (which is somewhat unfair to Pachi) is closer to meeting the assumptions underlying the selection MDP.

# 7 Conclusion

The selection problem has numerous applications. This paper formalized the problem as a belief-state MDP and proved some important properties of the resulting formalism. An application of the selection problem to control of sampling was examined, and the insights provided by properties of the MDP led to approximate solutions that improve the state of the art. This was shown in empirical evaluation both in "flat" selection and when extending the methods to game-tree search for the game of Go.

The methods proposed in the paper open up several new research directions. The first is a better approximate solution of the MDP, that should lead to even better flat sampling algorithms for selection. A more ambitious goal is extending the formalism to trees—in particular, achieving better sampling at non-root nodes, for which the purpose of sampling differs from that at the root.

# Acknowledgments

The research is partially supported by Israel Science Foundation grant 305/09, National Science Foundation grant IIS-0904672, DARPA DSO FA8650-11-1-7153, the Lynne and William Frankel Center for Computer Sciences, and by the Paul Ivanier Center for Robotics Research and Production Management.